\documentclass[twoside,leqno,twocolumn]{article}

\usepackage{amsmath,amssymb,amsfonts}
\usepackage{booktabs}
\usepackage{cite}
\usepackage{float}
\usepackage{graphicx}
\usepackage[letterpaper]{geometry}
\usepackage{hyperref}
\usepackage{SIAM/ltexpprt}
\usepackage{multirow}
\usepackage{subcaption}
\usepackage{textcomp}
\usepackage[normalem]{ulem}
\usepackage{xcolor}

\begin{document}

\newcommand\relatedversion{}
\renewcommand\relatedversion{\thanks{The full version of the paper can be accessed at \protect\url{https://arxiv.org/abs/1902.09310}}} 

\title{
    \Large Probabilistic Inverse Modeling: An Application in Hydrology
}


\author{
Somya Sharma\thanks{University of Minnesota - Twin Cities. \{sharm636, ghosh128, renga016, lixx5000, chatt019, nieber, kumar001\}@umn.edu, $^+$ Pennsylvania State University \{cxd11\}@psu.edu } \and
Rahul Ghosh${^*}$ \and
Arvind Renganathan${^*}$ \and
Xiang Li${^*}$ \and
Snigdhansu Chatterjee${^*}$ \and
John Nieber${^*}$ \and
Christopher Duffy${^+}$ \and
Vipin Kumar${^*}$
}

\date{}

\maketitle


\fancyfoot[R]{\scriptsize{Copyright \textcopyright\ 20XX by SIAM\\
Unauthorized reproduction of this article is prohibited}}

\pagenumbering{arabic}
\setcounter{page}{1}

\begin{abstract}

\small Rapid advancement in inverse modeling methods have brought into light their susceptibility to imperfect data.
The astounding success of these methods has made it imperative to obtain more explainable and trustworthy estimates from these models. In hydrology, basin characteristics can be noisy or missing, impacting streamflow prediction. For solving inverse problems in such applications, ensuring explainability is pivotal for tackling issues relating to data bias and large search space. We propose a probabilistic inverse model framework that can reconstruct robust hydrology basin characteristics from dynamic input weather driver and streamflow response data. We address two aspects of building more explainable inverse models, uncertainty estimation and robustness. This can help improve the trust of water managers, handling of noisy data and reduce costs. We propose uncertainty based learning method that offers 6\% improvement in $R^2$ for streamflow prediction (forward modeling) from inverse model inferred basin characteristic estimates, 17\% reduction in uncertainty (40\% in presence of noise) and 4\% higher coverage rate for basin characteristics.

\end{abstract}

\section{Introduction}
\label{sec:intro}

\begin{figure}
    \centering
    \includegraphics[width=0.5\linewidth]{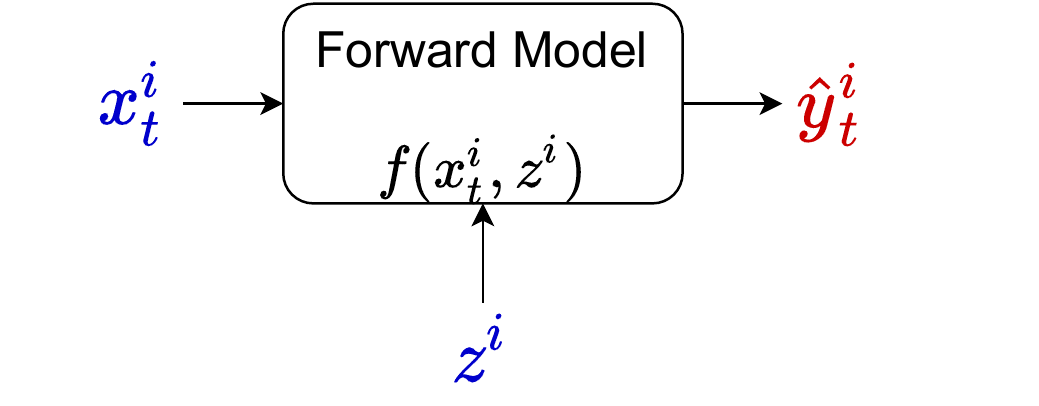}
    \caption{Forward model using weather drivers $x_i^t$ and river basin characteristics $z_i^t$ to estimate streamflow $y_i^t$ \cite{ghosh2021knowledge}}
    \label{fig:Forward}
\vspace{-15pt}
\end{figure}

Researchers in scientific communities study engineered or natural systems and their responses to external drivers. In hydrology, streamflow prediction is one crucial research problem for understanding hydrology cycles, flood mapping, water supply management, and other operational decisions. For a given entity (river-basin/catchment), the response (streamflow) is governed by external drivers (meteorological data) and complex physical processes specific to each entity (basin characteristics). Process-based models are commonly used to study streamflow in river basins (for example, Soil \& Water Assessment Tool). However, these hydrological models are constrained by assumptions, contain many parameters that need calibration and incur enormous computation cost. In addition, these models are often calibrated on every specific catchment and thus can require specific fine-tuning for each basin. As promising alternatives, machine learning (ML) models are increasingly being used ~\cite{kratzert2019towards} (Figure~\ref{fig:Forward} shows the diagrammatic representation of this data-driven forward model). In our study, an entity's response to external drivers depends on its inherent properties (called entity characteristics). For example, for the same amount of precipitation (external driver), two river basins (entities) will have very different streamflow (response) values depending on their land-cover type (entity characteristic)~\cite{newman2015gridded}. Disregarding these inherent properties of entities can lead to sub-optimal model performance. Knowledge-guided self-supervised learning (KGSSL) ~\cite{ghosh2021knowledge} has been proposed to extract these entity characteristics using the input drivers and output-response data. 

\begin{figure*}[t!]
    \centering
    \includegraphics[width=0.7\linewidth]{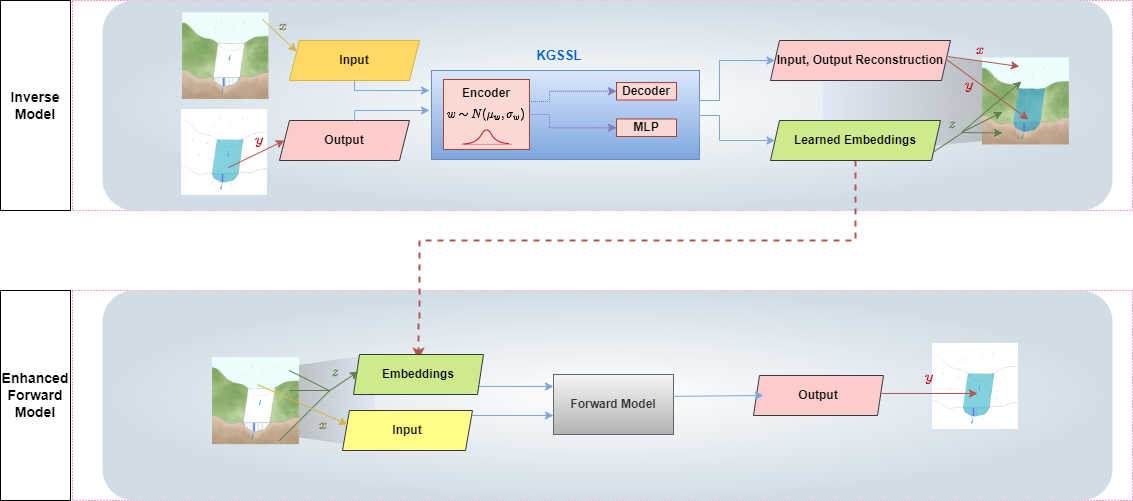}
    \caption{KGSSL for representation learning. The forward model learns streamflow ($y$) as functional approximation of weather drivers ($x$) and river basin static attributes ($z$). KGSSL leverages an inverse modeling framework for learning robust static attribute estimates ($\hat{z}$). The robust estimates aid in improving prediction performance of the forward model.}
    \label{fig:infographic}
    \vspace{-10pt}
\end{figure*}

Developing such entity-aware inverse models requires addressing several challenges. Often, the measured characteristics are only surrogate variables for the actual entity characteristics, leading to inconsistencies and high uncertainty. Uncertainty can arise due to several reasons, such as measurement error, missing data, and temporal changes in characteristics. Moreover, in real-world applications these characteristics may be essential in modeling the driver-response relation. However, they may be completely unknown, not well understood, or not present in the available set of entity characteristics. A principled method of managing this uncertainty due to imperfect data can contribute in improving trust of data-driven decision making from these methods. 

In this paper, we introduce uncertainty quantification in learning representations of static characteristics. Such a framework can help quantify the effect of multiple sources of uncertainty that introduce bias and error in decision-making. For instance, Equifinality of hydrological modeling (different model representation result in same model results) is a widely known phenomenon affecting the adoption of hydrology models in practice \cite{her2019uncertainty}. Uncertainty in model structure and input data are also widespread. In real world applications, studying these can help improve trust of water managers, improve hydrological process understanding, reduce costs and make predictions more explainable and robust \cite{mcmillan2018hydrological}. To achieve this, we propose a probabilistic inverse model for simultaneously learning representations of static characteristics and quantifying uncertainty in these predictions. As a consequence, we analyze the framework's reconstruction capabilities and its susceptibility to adversarial perturbations - our model is able to maintain same level of robustness as KGSSL. We also propose an uncertainty based learning (UBL) method to reduce epistemic uncertainty (uncertainty in predictions due to imperfect model and imperfect data) in our reconstructions. We show that it results in reducing the temporal artifacts in static characteristic predictions by 17\% and also reduces the epistemic uncertainty by 36\%. We also demonstrate the improvement in streamflow prediction (in the forward model) using these reconstructed static characteristics (6\% increase in test $R^2$). We provide model performance for reconstruction and forward modeling and compare it against the baselines, KGSSL \cite{ghosh2021knowledge} and CT-LSTM \cite{kratzert2019towards}, state-of-the-art for streamflow prediction. Since, we use a probabilistic model for estimating static characteristics, we obtain a posterior distribution instead of point estimates. This enables us to compute coverage rate of how often the observed values lie within the bounds of the inferred static characteristics' posterior prediction distribution. In practice, this can help water managers and public understand if we can reliably obtain a close enough prediction, even if we are not always accurate - analysis that can not be done with the deterministic inverse model. UBL offers a 4\% increase in coverage rate.

\vspace{10pt}

\section{Related Work}

\textit{Robustness:} In several large-scale applications in areas like computer vision and natural language processing, presence of even small, imperceptible adversarial perturbation can exacerbate model performance \cite{wei2019sparse, wallace2020imitation, eykholt2018robust}. While, several of the robustness studies focus on the effect of different noises \cite{li2018textbugger, zhou2019learning}, many other studies focus on methods to mitigate the adverse effects of these perturbations \cite{wang2019natural, shafahi2019adversarial, liu2020joint}. Through objective modification \cite{jia2019certified, katz2019marabou, fazlyab2020safety} or propagation of input-output relationship constraints  \cite{dvijotham2018training, huang2019achieving}, deep learning architectures have been modified to improve robustness. In real-world data sets, natural variations (like blurring) have been studied \cite{robey2020model, taori2020measuring}. Issues like adversarial transferability \cite{ilyas2019adversarial} and data brittleness issues (robustness issues due to overfitting \cite{rice2020overfitting}) bring to light the limitations of modern machine learning methods. More recent studies also look at Bayesian deep learning models for their robustness properties \cite{carbone2020robustness, cardelli2019statistical,sharma2021winsorization}. 

\textit{Inverse Problems:} In physical sciences \cite{ kim2018geophysical, woolway2021winter, pecha2021determination, dao2021improving}, several recent advances have focused on solving inverse problems. Unlike standard inversion methods in mathematics, that rely on non-linear optimization for calculating inverse of a forward model, recent machine learning methods allow us to learn the inverse mapping from datasets. This makes it imperative to mitigate any representation error and data biases before solving the inverse problem \cite{asim2020invertible}. Further, within these vast array of methodologies, selection of the right method is crucial - since, searching for an inverse mapping may be difficult due to the large search space. Bayesian optimization for searching and iterative gradient descent based methods may only provide a locally optimal inverse map \cite{lavin2021simulation}. Therefore, a principled MAP formulation or generative modeling may be viable for addressing these data related issues \cite{sun2020deep, whang2021composing, whang2021solving}. Inverse modeling approaches that rely on a single neural network may not accurately capture the spatio-temporal heterogeneity in basin characteristics. Moreover, entity characteristics can be unknown or noisy (due to measurement or estimation bias). Therefore, a robust and trust-worthy framework for learning entity characteristic can be useful in hydrology applications. Uncertainty quantification in inverse models also provides a responsible pathway for learning entity characteristic. We discuss this more in the next section.

\textit{Probabilistic Modeling:} Probabilistic models offer richer learning than a deterministic inverse model as they offer distributional recovery for the input embeddings. For a given input $x_i$, standard deviation of the prediction distribution is called epistemic uncertainty estimate ($\sigma_i$). Any deficiencies in the model framework (e.g., capturing only linear effects) and input data (e.g., missing / noisy data, low sample size) would increase this uncertainty. Smaller uncertainty estimates mean more ``confident" predictions. Several of the recent efforts use generative models for inverse problem solving \cite{whang2021composing, asim2020invertible, daw2021pid, whang2021solving}. 

We develop a Bayesian inverse model for robust recovery of complete distribution of the entity characteristics. Our framework achieves this by obtaining estimates of static variables from time series driver-response data. This, however, introduces temporal bias in our static characteristics. We propose an uncertainty based learning scheme to reduce the uncertainty associated with this temporal variation in inverse model estimates of static characteristics. We reconstruct static characteristics and use them instead of observed characteristics in forward modeling. 

\section{Method}
\label{sec:method}
\subsection{Autoencoder - based Inverse Model}
We use the CAMELS \cite{Addor2017} dataset with streamflow, weather driver and entity characteristic information (lake, river-basin or streams in river network). Each entity $i$ ($i=1,...,N$) has daily information, where $x_i^j \in \mathbb{R}^{\mathcal{D}_x}$ represents dynamic characteristics (weather drivers), $y_i^j \in \mathbb{R}^{\mathcal{D}_y}$ represents streamflow, $z_i^j \in \mathbb{R}^{\mathcal{D}_z}$ represents static characteristics for $i^{th}$ entity at $j^{th}$ time step. Similar to KGSSL \cite{ghosh2021knowledge}, our framework estimates static characteristics ($z$) from time-series information ($[x, y]$). Contrastive learning allows us to utilize the spatio-temporal correlation among river basins \cite{chen2020simple}. The objective function for training KGSSL is 

\begin{equation}
    \label{eq:representation learning}
    \small
    \mathcal{L} = \lambda_1 \mathcal{L}_{Rec} + \lambda_2 \mathcal{L}_{Cont} + \lambda_3 \mathcal{L}_{Inv}
\end{equation}

Where, reconstruction loss $\mathcal{L}_{Rec}$  enables accurate reconstruction of $[x, y]$; contrastive loss $\mathcal{L}_{Cont}$ utilizes the implicit relationships among driver-response time series data, enabling invariant approximation of static features; pseudo-inverse loss (or static loss) $\mathcal{L}_{Inv}$ utilizes available static variable examples to enable accurate representation learning. The loss weights are learned using hyper-parameter tuning. The Sequence Encoder, comprising of a bidirectional LSTM, encodes the driver-response time-series. Each (forward and backward) LSTM use $[\boldsymbol{x^t};y^t]$ input to generate carry state and the hidden state $h = [h_{\text{forward}}; h_{\text{backward}}]$.

Using a ReLU tranformation , a linear layer is used in the encoder to get a transformation of the hidden embedding. These transformed embeddings are used as input to the LSTM decoder $\mathcal{D}$. The observed sequence $\mathcal{S}_{e_i}$ are compared with the reconstructed sequence $\hat{\mathcal{S}}_{e_i}$ from the decoder in the reconstruction loss, $\mathcal{L}_{Rec} = \frac{1}{2N} \sum_{e\in \{a,p\}} \sum_{i=1}^N MSE(\hat{S}_{e_i}, S_{e_i})$.

\begin{equation}
\small
    \begin{split}
        \boldsymbol{i_t}    &= \sigma (\boldsymbol{W_i}\left[[\boldsymbol{x^t};y^t];\boldsymbol{h^{t-1}}\right] + \boldsymbol{b_i})\\
        \boldsymbol{f_t}    &= \sigma (\boldsymbol{W_f}\left[[\boldsymbol{x^t};y^t];\boldsymbol{h^{t-1}}\right] + \boldsymbol{b_f})\\
        \boldsymbol{g_t}    &= \sigma (\boldsymbol{W_g}\left[[\boldsymbol{x^t};y^t];\boldsymbol{h^{t-1}}\right] + \boldsymbol{b_g})\\
        \boldsymbol{o_t}    &= \sigma (\boldsymbol{W_o}\left[[\boldsymbol{x^t};y^t];\boldsymbol{h^{t-1}}\right] + \boldsymbol{b_o})\\
        \boldsymbol{c_t}    &= \boldsymbol{f_t} \odot \boldsymbol{c_{t-1}} + \boldsymbol{i} \odot \boldsymbol{g_t}\\
        \boldsymbol{h_t}    &= \boldsymbol{o_t} \odot \tanh{(\boldsymbol{c_t})}\\
    \end{split}
\end{equation}

Knowledge-guided Contrastive Loss ensures that the association among similar entities can allow for more efficient representation learning. The implicit physical properties (in embeddings $h_{a_i}$ and $h_{b_i}$) of ``positive pairs" of sequences ($S_{a_i}$ and $S_{p_i}$, respectively) are compared to other entity sequences.

\begin{equation}
\small
    \begin{split}
        l(a_i,p_i) = & \frac{\exp{(sim(\boldsymbol{h_{a_i}}, \boldsymbol{h_{p_i}})/\tau)}}{\sum_{e\in\{a,p\}}\sum_{j=1}^N\exp{(sim(\boldsymbol{h_{a_i}}, \boldsymbol{h_{e_j}})/\tau)}}\\
        + &\frac{\exp{(sim(\boldsymbol{h_{p_i}}, \boldsymbol{h_{a_i}})/\tau)}}{\sum_{e\in\{a,p\}}\sum_{j=1}^N\exp{(sim(\boldsymbol{h_{p_i}}, \boldsymbol{h_{e_j}})/\tau)}}
    \end{split}
\end{equation}
where, $sim(\boldsymbol{h_{a_i}}, \boldsymbol{h_{p_i}})=\frac{\boldsymbol{h_{a_i}}^T\boldsymbol{h_{p_i}}}{\|\boldsymbol{h_{a_i}}\|\|\boldsymbol{h_{p_i}}\|}$. Thus, the total contrastive loss for 2N such positive pairs is given as, $\mathcal{L}_{Cont} = \frac{1}{2N} \sum_{i=1}^N l(a_i,p_i)$.

$\mathcal{L}_{Cont}$ and $\mathcal{L}_{Rec}$ do not require any supervised information. This enables us to evaluate these losses on a large number of samples. Pseudo-Inverse Loss allows source of supervision based on the available static feature data. A feed-forward layer $I$ on sequence encoder output is used to estimate $\mathbf{\hat{z}} = I(\mathbf{h})$.

\begin{equation}
    \mathcal{L}_{Inv} = \frac{1}{N} \sum_{i=1}^N \frac{1}{z} \sum_{j=1}^z (z_i^j-\hat{z}_i^j)^2
\label{eq:Linv}
\end{equation}

Temporal heterogeneity in driver-response time-series is a source of uncertainty in the static feature reconstructions. For T time steps and W window size, $unc_i$ provides us with this standard deviation in static feature reconstruction over time, 


\begin{equation}
\label{eq:unc}
\small
    \boldsymbol{unc}_i = \sqrt{\frac{W}{T}\sum_{j=1}^{T/W} (\boldsymbol{\hat{z}}_i^j-\boldsymbol{\hat{z}}_i)^2}
\end{equation}

\subsection{Uncertainty Estimation}
\label{sec:method-uncest}

We modify the KGSSL framework to aid in uncertainty quantification. The uncertainty in estimation of static characteristics is obtained using a perturbation-based weight uncertainty method called \textit{Bayes by Backprop} \cite{blundell2015weight, wen2018flipout}. As a method that relies on learning posterior distribution of weight parameters, Bayes by Backprop makes different layers of the architecture non-deterministic. This allows us to measure and mitigate the uncertainty from different components incorporated in the framework.

Introducing perturbations in weights while training has historically been used as a regularization method \cite{hanson1988comparing, srivastava2014dropout, kang2016shakeout, li2016whiteout,goodfellow2013maxout}. Some recent advances utilize perturbations to induce non-deterministic behavior in supervised learning models \cite{graves2011practical, wan2013regularization}. Several variations of Bayesian neural networks implement the reparameterization trick \cite{kingma2015variational} to learn affine transformation of perturbation using variational inference. All these methods rely on drawing a Gaussian perturbation term $\epsilon \sim \mathcal{N}(0, 1)$. The scale and shift parameters $\Sigma$ and $\mu$ can be learned by optimizing for variational free energy \cite{graves2011practical}. Therefore, the weight parameters, $w$, are learned as, $w = \mu + log(1+exp(\Sigma)) \odot \epsilon$. Here, $log(1+exp(\Sigma))$ is non-negative and differentiable. The variational parameters $\theta = \{\mu, \Sigma \}$ are minimized by variational free energy \cite{graves2011practical, friston2007variational, jaakkola2000bayesian, yedidia2000generalized, neal1998view} that ensures a trade-off between learning a complex representation of the data (the likelihood cost) and learning a parsimonious representation similar to the prior (complexity cost). The variational free energy cost \cite{blundell2015weight} can be written as,

\vspace{-5pt}

\begin{equation}
    \mathcal{F} = KL[q(w| \theta)|| Pr(w)] - \mathbb{E}_{q(w| \theta)}[log \hspace{3pt} Pr(\mathcal{D}|w)]
\end{equation}

The complexity cost is the KL divergence between the learned posterior distribution of weight parameters $q(w| \theta)$ and the prior probabilitiy $Pr(w)$. The likelihood cost includes the negative log likelihood indicating the probability that the weight parameters capture the complexity of the dataset $\mathcal{D}$. Through this cost we are able to ensure that the weight distribution learns a rich representation and also does not overfit. The Gaussian perturbations in each mini-batch allows the gradient estimates of the cost to be unbiased.

In our sequence encoder, we obtain ReLU non-linear transformation of the final embeddings $h$ in a final linear layer. The weight parameters distribution in the linear layer are learned using Bayes by Backprop. We also tried other layers for learning parameter distribution.

\subsection{Uncertainty Based Learning (UBL)}

It is also imperative to manage uncertainty in complex deep learning architectures that may arise due to imperfect data. This can be achieved by penalizing static characteristics estimates with higher uncertainty. Uncertainty estimates from probabilistic models can therefore enable formulation of a regularization scheme to obtain lower uncertainty estimates. We can penalize the pseudo-inverse loss (Equation~\ref{eq:Linv}) such that the characteristics with higher uncertainty in the estimates will have higher loss due to bigger penalty coefficients.  In order to do that we look at the following theorem.

\begin{theorem}
Let $g$ be our inverse model receiving training set $S$ as input such that, $g_S: [x_t^i, y_t^i] \rightarrow z_i$. A loss $\mathcal{L}_g$ for prediction function $g_S$ defined as $\frac{1}{t} \frac{1}{N} \sum_{i=1}^N \frac{1}{|z|} \sum_{j=1}^z w^j (z_i^j-\hat{z}_i^j)^2$ minimizes uncertainty $\sigma_S$ where $w^j$ corresponds with $E_{\lambda_1}$, the eigenvector corresponding to the largest eigenvalue of $\sigma$.

\end{theorem}

Proof and more discussion given in Appendix. If our inverse model $g \in \mathcal{G}$, where $\mathcal{G}$ is a hypothesis class, the empirical risk of the hypothesis class, $ERM_{\mathcal{G}}$, can be decomposed into approximation and estimation error. While, our uncertainty guided methodology provides prior knowledge on the hypothesis class with lower uncertainty estimation for certain characteristics, the estimation error for other characteristics with lower penalty coefficients might be higher. As we will see in our experimental results, subset of features (geo-morphology based features) that were reconstructed successfully using KGSSL, saw an increase in bias in estimates from the UBL method. Also, the soil based features that were experiencing higher model approximation error in KGSSL estimates, saw an improvement in uncertainty estimates.

\begin{table}[]
    \tiny
    \centering
    \begin{tabular}{|p{3cm}|p{1cm}|p{1.4cm}|p{1.4cm}|}
    \hline 
    Average Metrics & Climate & Soil Geology & Geo-morphology \\
    \hline 
         Deterministic KGSSL RMSE & 0.292 & 0.575 & 0.430 \\
         Probabilistic KGSSL RMSE & 0.294 & 0.580 & 0.438 \\
         Deterministic KGSSL CORR & 0.958 & 0.792 & 0.878 \\
         Probabilistic KGSSL CORR & 0.958 & 0.788 & 0.873 \\
         Deterministic KGSSL UNC & 0.182 & 0.290 & 0.218 \\
         Probabilistic KGSSL UNC & 0.175 & 0.258 & 0.205 \\
         \hline 
    \end{tabular}
    \caption{Model performance of deterministic and probabilistic KGGSL in terms of test RMSE, test correlation between predicted and observed static feature and test uncertainty over time, $unc$. The metrics are averaged over 9 features in each category. Probabilistic KGSSL is able to achieve same level of model performance and allows for uncertainty quantification as well. Robust estimates from probabilistic model lead to slightly lower uncertainty over time values.}
    \label{tab:static_reconstruction_detvsprob}
\end{table}

\section{Results}
\label{sec:results}

\textbf{Dataset:} We use the CAMELS dataset, which is a publicly available hydrology dataset for multiple hydrology entities (including the 531 entities that were included in our study). The input variables for the forward model are 5 time-varying weather drivers and 27 static characteristics about the entities (weather descriptive statistics, soil based features, and geo-morphology based features are included in the study. These affect the streamflow, water runoff processes). The response variable is streamflow values. In practical setting, static characteristics information for all the entities may not be known. This makes it imperative to explore representation learning frameworks that can provide inferred characteristics for predicting streamflow for all entities. In our inverse model, the streamflow - weather time series are used for learning static characteristics.

\textbf{Experimental Setup:} Daily data from year 1980 - 2000 are used for training, year 2000 - 2005 are used for validation and year 2005 - 2015 are used for testing.  We report mean squared error (MSE), NSE (Nash-Sutcliff Efficiency is a measure similar to $R^2$. It is used to measure prediction performance in time-series hydrological models), and uncertainty estimates (standard error in prediction estimates). We predict static characteristics for all 531 river basins in the test period. These predictions are made from KGSSL and probabilistic encoder based KGSSL.

\subsection{Epistemic Uncertainty}

The probabilistic inverse model can be obtained by making different components of the framework non-deterministic as suggested in Sub-section 3.2. A probabilistic encoder based inverse model not only provides us with best validation based inverse loss, but also allows uncertainty quantification in basin characteristics reconstructions and driver-response reconstructions. More details on exploring different sources of uncertainty in appendix. Moreover, uncertainty over time, $unc_i$, and epistemic uncertainty, $\sigma_i$, have been shown to be correlated. This association enables a loss formulation where we can penalize reconstructions resulting in higher epistemic uncertainty to obtain a learned model where uncertainty over time is also reduced. More discussion provided in appendix.

\begin{figure}[h]
    \centering

    \includegraphics[width=0.5\textwidth]{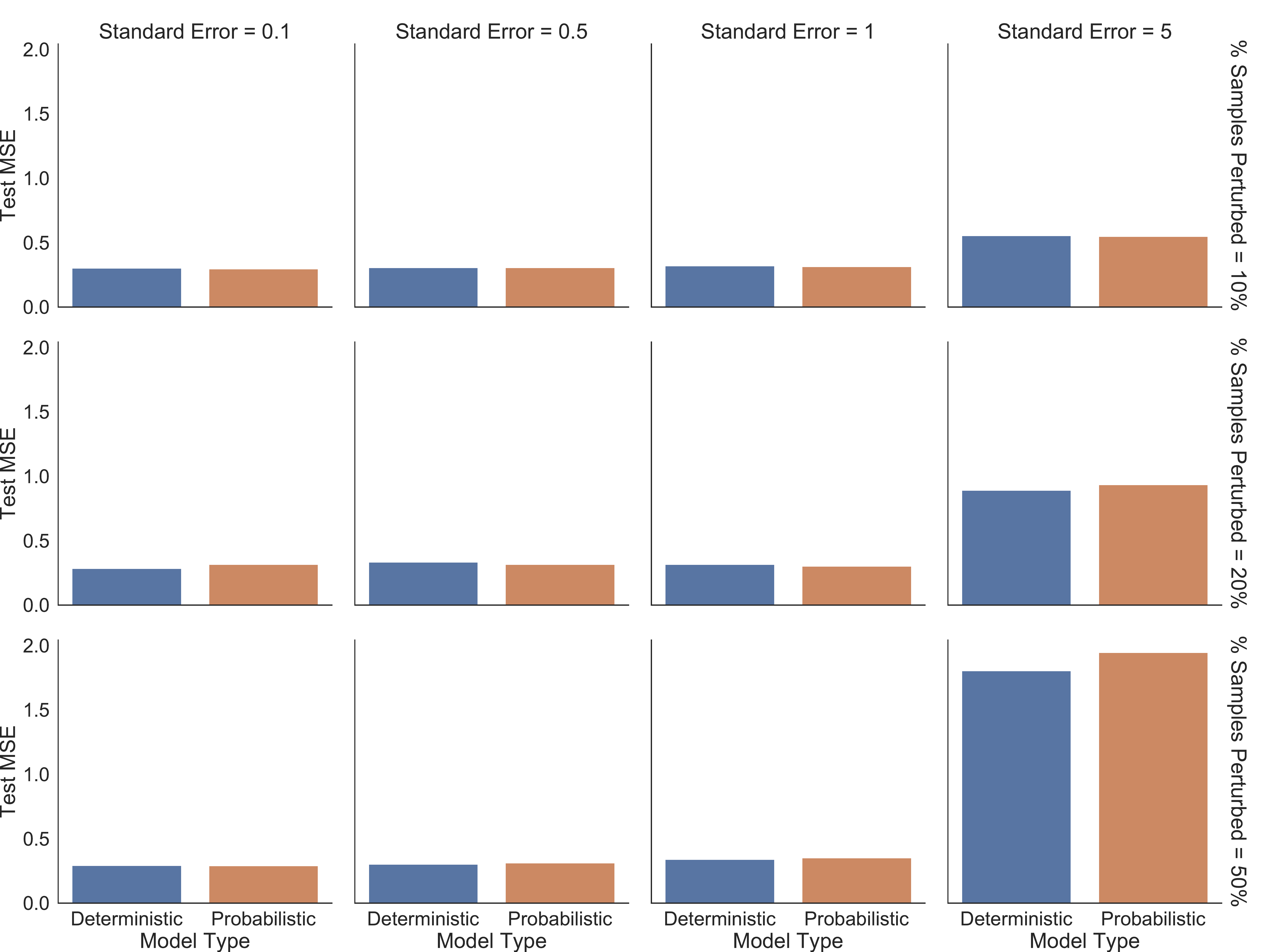}
    \caption{Robustness to noise with $10\%$ samples perturbed, $20\%$ samples perturbed and $50\%$ samples perturbed at 4 levels of standard error. The probabilistic model is able to maintain same level of bias as the deterministic model.}
    \label{fig:robustness}
    \vspace{-10pt}
\end{figure}

\begin{figure}[h]

    \centering
    \includegraphics[width=0.5\textwidth]{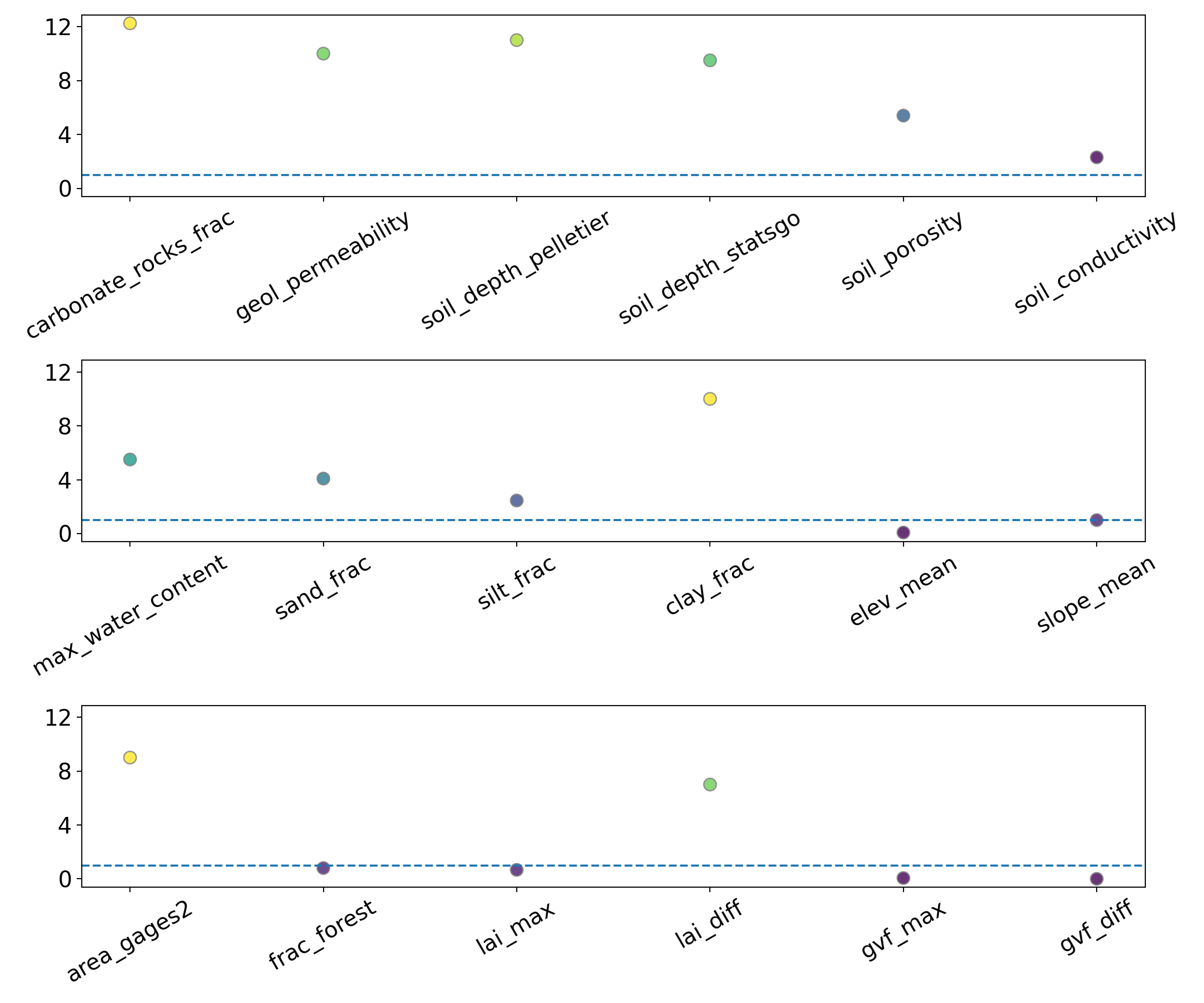}
\caption{Trade-off ratio is computed as ratio of fraction decrease in uncertainty estimates and fraction increase in MSE values using the Uncertainty Based Learning.}
\label{fig:ub_tradeoffratio}

\vspace{-10pt}
\end{figure}

    

\subsection{Robustness}

Due to the imperfection in driver-response data, it is important to address robustness of our framework to noise. We evaluate the static characteristics reconstruction by perturbing different fractions of training data with different size of perturbations.  Table~\ref{tab:robustness} shows the reconstruction performance for weather drivers and streamflow as \textit{reconstruction MSE loss} and reconstruction performance for static characteristics as the \textit{static loss}. Since accurate reconstruction of static variables is of interest, the epistemic uncertainty in the static estimates are mentioned as \textit{static uncertainty}. The percent of training examples that are perturbed with noise is varied from 1\% to 50\% while the perturbation size is determined by the standard error of the Gaussian noise that is added to the trainind data. Training examples are perturbed and model performance is evaluated on the clean test set. The static reconstruction loss increases marginally with the addition of noise when the standard error remains below 5. In Table~\ref{tab:robustness} and Figure~\ref{fig:robustness}, for deterministic and probabilistic model, we notice similar levels of bias in static feature reconstruction, where the probabilistic model can also measure the variability in the prediction estimates.

\begin{table*}[]
\tiny
    \centering
    \begin{tabular}{|c|c|c|c|c|}
        \toprule
        \textbf{Method}  &  \textbf{Noise}  & \textbf{Reconstruction Loss}     & \textbf{Static Loss} & \textbf{Static Uncertainty} \\
        \midrule
        Deterministic   &   -   &   \textbf{1.940}  &  0.3137    & 0.000              \\
        \hline
        Probabilistic Encoder    &   -   &   2.022  &  \textbf{0.2829}   &  0.0011  \\
        \hline
        Deterministic   &   1\% corruption, 10 s.d. & 1.937  &   0.3001  & 0.000\\
        \hline
        Probabilistic Encoder    &   1\% corruption, 10 s.d. &   1.935 & 0.3119    & 0.0017 \\
        \hline
        Deterministic   &   5\% corruption, 10 s.d. & 1.947 & 0.3095    &   0.000\\
        \hline
        Probabilistic Encoder    &   5\% corruption, 10 s.d. &   2.004 &   0.3054  &   0.0021 \\
        \hline
        Deterministic (UBL)   &   1\% corruption, 10 s.d. & 1.955  &   0.3021  & 0.000\\
        \hline
        Probabilistic Encoder (UBL)    &   1\% corruption, 10 s.d. &   1.938 & 0.3135    & 0.0008 \\
        \hline
        Deterministic (UBL)   &   5\% corruption, 10 s.d. & 1.944 & 0.3128    &   0.000\\
        \hline
        Probabilistic Encoder (UBL)    &   5\% corruption, 10 s.d. &   1.946 &   0.3234  &   \textbf{0.0007} \\

        \bottomrule
    \end{tabular}
    \caption{\small Streamflow, weather driver reconstruction MSE test loss, static characteristic MSE test loss and static characteristic estimate epistemic uncertainty. The model performance in terms of test MSE for higher levels of noise in training data are given in Figure~\ref{fig:robustness}. We can also look at robustness results for UBL variants - the bias increases slightly while the epistemic uncertainty decreases marginally.} 
    \label{tab:robustness}
  
\end{table*}



The bias in prediction estimates are robust to corruption in training data, but the variance in estimates increases. This may also impact the association that we observe between the uncertainty over years and epistemic uncertainty. Figure 4 shows us the correlation between the two uncertainties. To observe the effect of higher levels of perturbation, we look at the scenario when 50\% of the training data is perturbed.In sub-figure 4b, we notice increased correlation between the weather and geo-morphology based static features. The increased epistemic uncertainty due to perturbations, indicate a greater variability in static features, over time and otherwise. Therefore, with increased variance in static feature reconstruction, the correlation between the two types of uncertainty (measuring the deficiency in our input data), also increases with higher perturbation levels.

\subsection{Uncertainty Based Learning (UBL)}
\label{sec:results-ubl}

Learning static characteristics using the UBL methodology, we penalize reconstructed representations with higher uncertainty estimates in preliminary modeling performance (on validation dataset). Allowing for regularization that manages uncertainty results in a model that has lower uncertainty estimates and lower variance in static features over time. This however, also leads to increase in bias for all static features. We can look at the percent decrease in uncertainty estimates and percent increase in mean squared error values on the test set to understand the trade off between variance and bias. Figure~\ref{fig:ub_tradeoffratio} shows us the trade off ratio computed as the ratio of percent decrease in uncertainty and percent increase in test set MSE values. The blue dashed line represents a trade-off ratio of 1. Trade-off ratio above 1 represent scenarios where percent improvement in uncertainty is more than the percent increment in MSE values for a given variable. For instance, for "carbonate rocks fraction" variable, the percent decrease in uncertainty was 25\% while the increase in MSE was 2\% - therefore, the trade-off ratio is 12.5. We notice, for the soil based static features, UBL resulted in improvement in uncertainty. Since, geo-morphology variables like ``elevation mean" and ``slope mean" are already well approximated with lower temporal variance in the estimates, the trade-off ratios reflect the ineffectiveness of UBL here.

We can also compare the UBL model performance in terms of test set NSE values with the deterministic and probabilistic KGSSL model (Table~\ref{tab:inverse}). In addition to improvement in static feature reconstruction, the probabilistic model also enable us to measure if our mean prediction estimates lie withing the confidence intervals created using the standard error in the predictions. We look at the one standard deviation and two standard deviation confidence interval coverage rates.

\vspace{-5pt}

\begin{equation}
\small
    \text{coverage rate} = \frac{\mathbb{I} (z_i \in [\mu_{z_i} - \sigma_{z_i}, \mu_{z_i} + \sigma_{z_i}])}{N \times |z|}
    \vspace{-5pt}
\end{equation}

While we obtain a lower test NSE using the UBL method, we are able to obtain a higher coverage rate than the probabilistic KGSSL method. This may indicate that the regularization of uncertain predictions result in more confident estimation with wider coverage. In Table~\ref{tab:robustness}, we also compare the bias and uncertainty in the presence of noise in training examples. We notice similar levels of robustness as compared to the previous models.

\begin{table}[h]
\tiny
    \centering
    \begin{tabular}{|p{3cm}|c|p{1.2cm}|p{1.2cm}|}
    \toprule
          Model  & NSE & 63\% C.I. Coverage Rate & 95\% C.I. Coverage Rate\\
          \hline
          KGSSL & 0.6556 & - & - \\
          Probabilistic KGSSL  & \textbf{0.6858} & 0.8169 & 0.9386 \\
          KGSSL (UBL) & 0.6587 & - & -  \\
          Probabilistic KGSSL (UBL) & 0.6669 & \textbf{0.8220} & \textbf{0.9783}\\
          \bottomrule
    \end{tabular}
    \caption{Static reconstruction NSE and coverage rate. We can compare the static characteristic reconstruction NSE values among the deterministic and probabilistic models. Probabilistic models also ensure that our predictions will lie within the (mean $\pm z_{\alpha}$  s.d) interval.}
    \label{tab:inverse}
\end{table}

\begin{table}[h]
    \tiny
    \centering
    \begin{tabular}{|p{3cm}|c|c|}
    \toprule
          Model  & Average NSE & Ensemble NSE\\
          \hline
          Baselines EALSTM (original static characteristics)  & 0.7031 & 0.7238 \\
          KGSSL & 0.7501 & 0.7570 \\
          Probabilistic KGSSL  & 0.7561 & \textbf{0.7597} \\
          KGSSL (UBL) & 0.7611 & 0.7582 \\
          Probabilistic KGSSL (UBL) & \textbf{0.7636} & 0.7594\\
          \bottomrule
    \end{tabular}
    \caption{NSE in forward model streamflow prediction using reconstructed static characteristics as input. Over 5 runs, we build 5 inverse and forward models. Average NSE is average of test NSEs obtained from each forward model. Ensemble NSE is computed from average of predictions from the 5 runs.}
    \label{tab:forward}
\end{table}
\subsection{Enhanced Forward Model}

In our analysis, we are investigating methods to reconstruct static characteristic values that are not only more robust to noise but also enables estimation in scenarios where samples might be missing in the observed data. KGSSL enables representation learning that solves the missing-ness and robustness challenges in the real world static characteristic data \cite{ghosh2021knowledge}. We compare forward modeling results in streamflow prediction using original and reconstructed static characteristic values. As can be seen in Table~\ref{tab:forward}, KGSSL offers improvement in forward model performance using reconstructed static features. We are also able to see improvement using the probabilistic KGSSL model and the UBL variants.

\section{Conclusion}
\label{sec:conclusion}
Rapid advancements in deep learning methods have made it imperative to obtain more explainable and trust-worthy decisions from these models. Even more so, for solving inverse problem, ensuring explainability is critical. In hydrology, a probabilistic inverse model offers us the ability to infer basin characteristics that are more trust-worthy. This eliminates the need for thorough curating of large datasets that might be very expensive and time-taking \cite{gebru2017fine}. This method offers us improvement in streamflow prediction skill, offers the same level of robustness as previous methods and provides a wider coverage rate as well. We also incorporate inductive bias through our prior assumption on behavior of static characteristics. While this method utilizes penalty coefficients that were estimated in ``one-shot" on the validation dataset, in the future work a more adaptive approach can be explored to learn the penalty coefficients during training. Similar to a learnable dropout rate \cite{gal2016dropout, gal2017concrete}, such a method may utilize a variational energy optimization scheme to learn these coefficients. 

In hydrology, probabilistic inverse modeling can offer many insights. Better reconstructions for variables like soil porosity and conductivity imply their impact on streamflow generation process is easily predictable as they govern soil water permeation behavior more closely. Variables like carbonate rock fraction signify differences in topology (Karst geology) which are not as easily predictable - this is also showcased in lower prediction skill of the inverse model and higher uncertainty. Therefore, decision makers can be more cautious about inferred basin characteristics that have higher uncertainty.

Beyond the scientific applications of streamflow in river basins, the need to build robust and personalized prediction models exists in several real-world applications. For example, while predicting mood (response), the personality (characteristics) of the patient (entity) will impact how the patient responds to outside weather (driver), or while assessing the aesthetics of an image (driver), the rating (response) by the assessor (entity) depends on the personality (characteristics). Thus, our framework can be generalized to setting where taking into account these entity characteristics will be essential.

\bibliographystyle{plain}
\bibliography{references}

\end{document}